\documentclass[10pt,twocolumn,letterpaper]{article}

\usepackage{cvpr}
\usepackage{times}
\usepackage{epsfig}
\usepackage{graphicx}
\usepackage{amsmath}
\usepackage{amssymb}

\usepackage[breaklinks=true,bookmarks=false]{hyperref}

\cvprfinalcopy 

\def\cvprPaperID{****} 

\begin{document}

\title{NeuroTreeNet: A New Method to Explore Horizontal Expansion Network}

\author{Shenlong Lou$^1$, Yan Luo$^1$, Qiancong Fan$^1$, Feng Chen$^1$, Yiping Chen$^1$, Cheng Wang$^1$, Jonathan Li$^1$$^*$\\ \vspace{5mm}
$^1$Fujian Key Laboratory of Sensing and Computing for Smart Cities, Xiamen University, China\\
{\tt\small loushenlong@stu.xmu.edu.cn, junli@xmu.edu.cn}
}

\maketitle

\begin{abstract}
It is widely recognized that the deeper networks or networks with more feature maps have better performance. Existing studies mainly focus on extending the network depth and increasing the feature maps of networks. At the same time, horizontal expansion network (e.g. Inception Model) as an alternative way to improve network performance has not been fully investigated. Accordingly, we proposed NeuroTreeNet (NTN), as a new horizontal extension network through the combination of random forest and Inception Model. Based on the tree structure, in which each branch represents a network and the root node features are shared to child nodes, network parameters are effectively reduced. By combining all features of leaf nodes, even less feature maps achieved better performance. In addition, the relationship between tree structure and the performance of NTN was investigated in depth. Comparing to other networks (e.g. VDSR\_5) with equal magnitude parameters, our model showed preferable performance in super resolution reconstruction task.
\end{abstract}

\section{Introduction}

\begin{figure}\small
  \centering
  \includegraphics[width=0.48\textwidth, height=0.47\textheight]{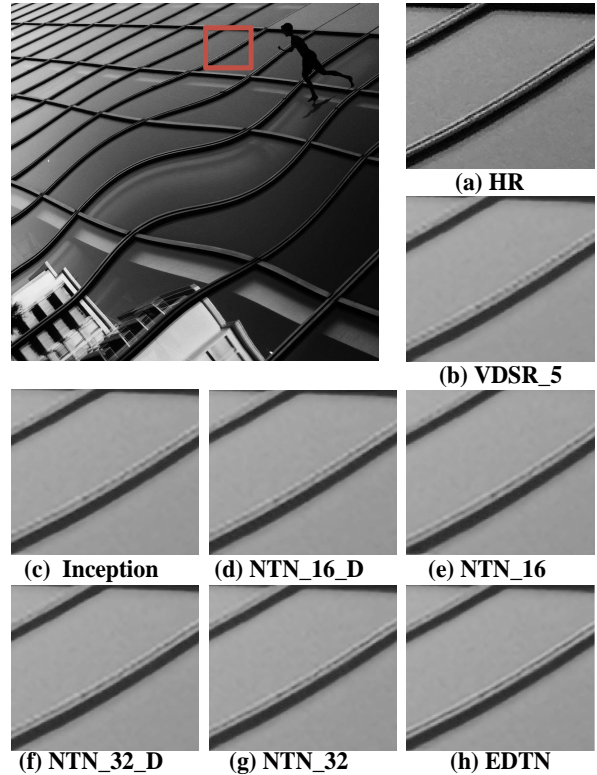}\\
  \caption{Super-resolution result on 3¡Á enlargement. PSNR: VDSR\_5 \cite{kim2016accurate} (33.20 dB), Inception \cite{szegedy2015going} (33.37 dB), NTN\_16\_D (33.45 dB), NTN\_16 (33.61 dB), NTN\_32\_D (33.64), NTN\_32 (33.91), and EDTN (34.06 dB).}
  \label{fig1}
\end{figure}

Research in deep learning has made significant progress in recent years, especially for convolutional neural networ-ks (CNNs), which have produced many classic models (e.g. AlexNet) \cite{krizhevsky2012imagenet,he2016deep,huang2017densely,szegedy2015going,simonyan2014very,arbelaez2011contour}. In contrast to traditional machine-learning, CNNs based models don¡¯t need features predesigned artificially but extract features directly from data due to its strong feature learning ability. CNNs have shown great superiority in image classification, object detection and voice recognition in recent years \cite{krizhevsky2012imagenet,hinton2012deep,girshick2014rich}.

Since the emergence of residual learning \cite{he2016deep}, there is a consensus that deeper networks have better performance. Residual Network (ResNet) solves the gradient dispersion problem \cite{he2016deep}, which leads researchers to focus on deepening networks \cite{kim2016accurate,kim2016deeply,lim2017enhanced,tai2017image,szegedy2017inception,zhang2017beyond,yu2017automated}. Although the deeper network showed superiority in various tasks \cite{lim2017enhanced,zhang2017beyond,yu2017automated}, a major problem that comes with it is the bloat of the model. It is still a question worth considering about the tradeoff between performance and model bloat.
\begin{figure*}[t]\small
  \centering
  \includegraphics[width=0.92\textwidth, height=0.44\textheight]{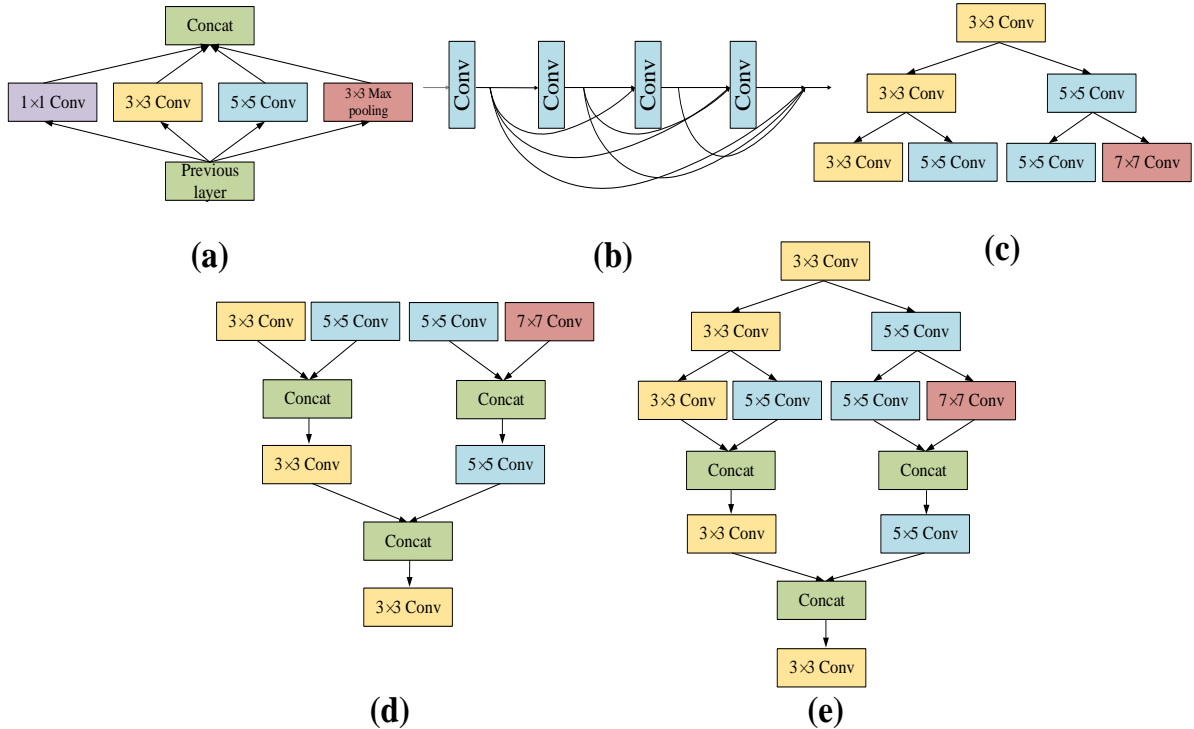}\\
  \caption{Different feature fusion models. (a) Inception model consists of three convolutional filters of different sizes and one max pooling which shows superiority in image classification and object recognition. (b) DenseNet, the input of each layer includes the output of all the previous layers of the network. (c) Neural Tree Net, following the structure of the binary tree which each layer consists of several convolutional filters of different sizes. (d) Reverse Tree Net, the inverse form of neural tree network. (e) Encoder-decoder Tree Net, combing Neural Tree Net and Reverse Tree Net according to the ideology of auto-encoder network}
  \label{fig2}
\end{figure*}

Besides expanding the depth of the network or increasi- ng the feature maps of networks \cite{he2016deep,huang2017densely}, another way to make the network obtains better capability is extending the network horizontally (e.g. Inception Model). Each layer of a Horizontal Expansion Network (HEN) consists of multi-scale convolution filters. HEN is more suitable for solving multi-scale problem: small object detection, image inpain-ting and super resolution reconstruction (SR), because these tasks require multi-scale information capture capability of the model. Compared to traditional methods, which scale data to get multi-scale information \cite{nah2017deep}, HEN obtains multi-scale information by using multi-scale convolution operators. However, according to our understanding, the research of HEN is still at limited.

In this paper, we proposed a new HEN named Neuro- TreeNet (NTN), as inspired by Random Forest (RF) \cite{liaw2002classification} and Inception Model (a typical HEN) \cite{szegedy2015going}. The core content of NTN is tree structure, which means other networks can convert to NTN by adding tree structure into their architecture. Our networks achieved better results in super resoluteion reconstruction, compared with other lightweight networks, as illustrated in Fig. \ref{fig1}. The main contributions of our work are summarized as follows:

\textbf{(1) Tree structure.} We proposed a new method to explore horizontal expansion network, in which the ideology of tree was included. The special structure makes full use of multi-scale information.

\textbf{(2) Feature fusion and shared feature.} Thanks to the special structure of tree which makes the feature fusion more flexible compared to other HENs. As Fig. \ref{fig2} (c) and (d) shows, each branch of the tree can be thought of as an independent neural network, which means we could adjust the richness of features in the final feature fusion by controlling the number of nodes in each layer. Another advantage is that child nodes share the features of the root node, which is a good strategy for reducing parameters.

\textbf{(3) Research on the attributes of NTN network.} In this paper, the number of nodes and the location of the tree in the network and their impact on performance are explored. Meanwhile, a variable is defined to quantitatively describe the contribution of each branch to tree model. In addition, we also discuss two kinds of new combination of tree model.

\section{Related Work}
\subsection{Horizontal expansion network}
In recent years, research on neural networks has been mainly focused on expanding the depth of model and number of features \cite{he2016deep,huang2017densely,kim2016accurate,kim2016deeply,lim2017enhanced,tai2017image,szegedy2017inception,zhang2017beyond,yu2017automated}, but neglecting the investigation of its breadth. Up to now, only Inception model is truly a HEN, of which each layer consists of multiple convolution filters with multi-scale information, as shows in Fig. \ref{fig2} (a). Aiming at the problem of parameter proliferation caused by expansion, the corresponding solutions are proposed in several subsequent Inception variant models \cite{ioffe2015batch,szegedy2016rethinking,szegedy2017inception}. Although the GoogleNet consisting of multiple inception models has achieved good results in various fields such as image classification and object recognition \cite{szegedy2015going}, there are still some issues with this model such as scale solid-fication and inflexible. In addition, the relationship between network width and model performance of the GoogleNet has not been investigated.

\subsection{Feature fusion methods}

For some special tasks (i.e. small object detection), the use of multi-scale information is critical to the result. Meanwhile, feature fusion is an inevitable process using multi-scale features \cite{nah2017deep}.

The main feature fusion methods in the field of deep learning are divided into two categories: multi-network feature fusion \cite{bodla2017deep,szegedy2015going,jeong2017enhancement} and single network feature fusion \cite{huang2017densely,bodla2017deep}. Multi-network feature fusion is always accompanied by multi-scale features, like Inception Model \cite{szegedy2015going} or a set of multiple independent networks \cite{bodla2017deep}. Compare to multi-network feature fusion, the single network feature fusion is mainly used for the fusion of low-dimensional features and high-dimensional features, including only single-scale information, like Dense Network (DenseNet) \cite{huang2017densely}.

As shown in Fig. \ref{fig2} (b), the DenseNet stacks the features of the front layer to the following layers by depth, the features after stacking will be mixed by convolution filter. The unique architecture of DenseNet takes full advantage of each layer¡¯s features (both low-level and high-level),accompanied by significant performance improvements. However, two major flaws of DenseNet are: (1) feature redundancy, repeatedly stacking front layer features to the following layers; (2) single-scale, not enough to deal with multi-scale tasks.

\begin{figure}\small
  \centering
  \includegraphics[width=0.45\textwidth, height=0.12\textheight]{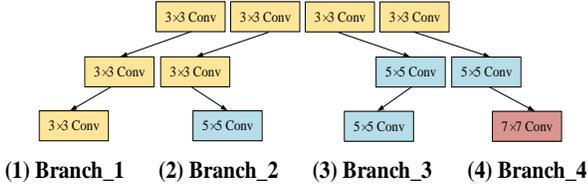}\\
  \caption{ Branch network of NTN network.}
  \label{fig3}
\end{figure}

\subsection{Deep learning for image super resolution}

The deep learning method has gradually occupied a dominant position in the field of SR due to its powerful feature extraction ability \cite{dong2014learning,kim2016accurate,kim2016deeply,lim2017enhanced,tai2017image,timofte2014a+,dong2016accelerating}.

Since SRCNN \cite{dong2014learning} defeated other machine learning methods in SR, more innovative networks have been proposed to improve the results of super resolution reconstruction. Compared to SRCNN with only three layers and 64 feature maps in each layer, the subsequent network either extended the depth (e.g. VDSR \cite{kim2016accurate}) or raised the number of feature maps (e.g. DRCN \cite{kim2016deeply}), or even both (e.g. EDSR \cite{lim2017enhanced}). It is worth recognizing that the reconstruction results have indeed improved to a certain extent, which also confirms that the deeper network or network with more feature maps has better performance. However, none of these models use multi-scale information and consider horizontal expansion.

We applied our NTN in SR, the VDSR that a type single-scale network was chosen as our baseline due to its simple structure and small amount of computation. The application of multi-scale convolution and feature fusion makes our network have stronger feature extraction ability. Compared with VDSR, our model achieves better results in all scales with less parameters.

\section{Neural Tree Network}

We proposed the NTN which introduces the ideology of Random Forest \cite{liaw2002classification}, each neural network can be regarded as a tree due to the existence of hidden layer units. Some common problems in HENs, such as scale solidification and inflexibility were solved by this special structure.

\subsection{Tree structure}

Tree structure is a well-known data storage structure, and it is also the core module of RF \cite{liaw2002classification} and Decision Tree (DT) \cite{brodley1995multivariate}. The architecture of tree we proposed is more like binary tree that each parent node has no more than two child nodes. Considering the limitation of the parameter quantity, other tree structures are not explored in this paper. We define that the convolutional size for every node as follows:
\begin{equation}
S_N^l = S_{N-1}+n
\label{eq2}
\end{equation}

\begin{equation}
S_N^r = S_{N-1}
\label{eq2}
\end{equation}
where $S_N^l$ and $S_N^r$ are left child node convolutional size and right child node convolution size at layer N respectively, $n$ is a constant number which we set to 2 in this work. These definitions ensure that each branch of NTN is unique, making NTN more diverse.

Different with single-scale networks, where each layer performs a convolution operation, our NTN represents a convolution operation at each node. As shown in Fig. \ref{fig3}, an NTN can be regarded as a set of deep neural networks (DNNs):

\begin{figure*}
	\centering
	\includegraphics[width=0.87\textwidth, height=0.30\textheight]{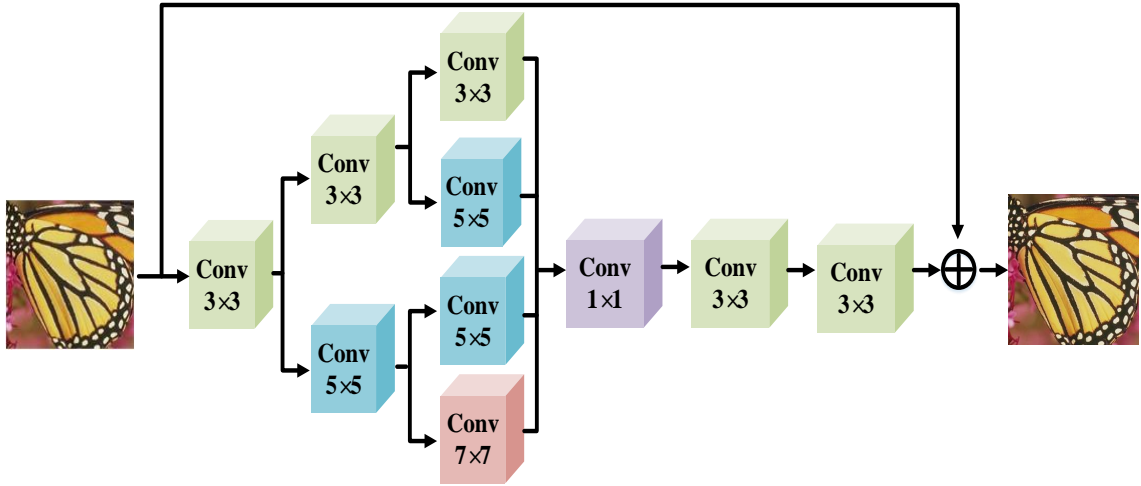}\\
	\caption{NTN architecture for SR.}
	\label{fig4}
\end{figure*}

\begin{equation}
NTN=\{DNN_1,DNN_2,...,DNN_t\}
\label{eq2}
\end{equation}
where $t$ means the number of the leaf nodes. NTN can be integrated with different levels of DNNs for different difficult tasks. Meanwhile, training an NTN is equivalent to training several DNNs at same time, which will greatly improve efficiency in some tasks.

Due to the existence of multiple branches, the final feature fusion for all different scales is inevitable. Different from the single-scale network feature fusion, the feature fusion of NTN is based on multi-scale network, which greatly increase the richness of features. Meanwhile, NTN obtains different degrees of multi-scale features by adjusting the convolutional size of nodes, which is more flexible than Inception Model. Another advantage is that each branch of NTN learns less features but get better performance after fusing all features of branches.

In addition, NTN has a unique advantage is feature sharing that solves problems in parameter explosion and feature redundancy. The feature sharing in NTN is defined as follows:
\begin{equation}
F_N^r = F_N^l=F_{N-1}*W_{N-1}
\label{eq2}
\end{equation}
\begin{equation}
F_{N+1}^r=F_N^r*W_N^r
\label{eq2}
\end{equation}
\begin{equation}
F_{N+1}^l=F_N^l*W_N^l
\label{eq2}
\end{equation}
where $F_N^r$ and $F_N^l$ are input features of left nodes and right nodes at layer $N$ respectively, $W$ means the weight of convolutional filters.

To verify the performance of our models, we extend the tree structure to NTN and apply it in SR.

\subsection{Branch contribution}
\label{Branch}
As mentioned in previous sections, each branch of the NTN network can be regarded as an independent network. The branch network shares features through nodes, and each network extracts different information from input data. Thus, each branch network has different contribution to the entire network.

Contribution of each branch of NTN in feature fusion was measured through three indicators including PSNR value, parameter quantity, and the improvement of the receptive field brought by the branch. Furthermore, to better understand the quantitative relationship among above indicators, the contribution index (CI) defined as follows was used.

\begin{equation}
CI_i=\frac{\beta_iPVI_i+\lambda RFI_i}{e^{PQI_i}}(i=1,2,3\cdots)
\label{eq2}
\end{equation}
\begin{equation}
\beta_i=\frac{P_0-min(P_i)}{P_0-P_i}(i=1,2,3\cdots)
\label{eq2}
\end{equation}

\begin{equation}
PVI_i=\frac{P_i-[P_i]}{P_0-[P_i]}(i=1,2,3\cdots)
\label{eq2}
\end{equation}

\begin{equation}
RFI_i=\frac{R_i}{R_0}(i=1,2,3\cdots)
\label{eq2}
\end{equation}

\begin{equation}
PQI_i=\frac{Pq_i}{Pq_0}(i=1,2,3\cdots)
\label{eq2}
\end{equation}
where PVI, RFI, PQI are the variables to describe PSNR value, receptive field, parameter quantity of each branch network, respectively. $P_0$, $R_0$, $Pq_0$ are the whole NTN¡¯s' PSNR value, receptive field, parameter quantity, respectively. $P_i$, $R_i$, $Pq_i$ are each branch of NTN¡¯s' PSNR value, receptive field, parameter quantity and we define ¡°[]¡± as a rounding down operation. $\beta_i$  is a parameter for weighting PVI. We believe that when the performance of the branch network is close to the NTN network, it is more difficult to have further promotion, so we set parameter $\beta_i$ as the ¡°reward¡±. $\lambda$ is a hyper-parameter to weighted RFI, we consider that the PSNR value is of higher importance in the evaluation task, so we set $\lambda$ value to 0.5.

\subsection{Network architecture for SR}

\begin{table*}\small
	\caption{The average PSNR/SSIM for scale factor ¡Á2, ¡Á3 and ¡Á4 on datasets Set5, Set14, B100 and Urban 100. {\color{red}{ Red}} color indicates the best performance and {\color{blue}{ blue }}color indicates the second best performance.}
	\begin{tabular}{|c|c|c|c|c|c|c|c|c|}
		
		\hline
		Dataset &  Scale &  VDSR\_5 & Inception & NTN\_16\_D & NTN\_16 & NTN\_32\_D &NTN\_32 &EDTN \\
		\hline  & ¡Á2 & 36.69/0.9554 & 36.72/0.9549 & 36.74/0.9546 & 36.85/0.9552 & 36.86/0.9552 &{\color{blue}{ 36.97/0.9558}} &{\color{red}{ 37.08/0.9563}} \\
		Set5 & ¡Á3 & 32.73/0.9085 & 32.88/0.9160 & 32.91/0.9105 & 32.99/0.9116 & 33.05/0.9121 &{\color{blue}{  33.09}}/{\color{red}{0.9219}} & {\color{red}{33.27}}/{\color{blue}{0.9148}} \\
		& ¡Á4 & 30.47/0.8627 & 30.61/0.8660 & 30.62/0.8659 & 30.67/0.8672 & 30.74/0.8659 &{\color{blue}{  30.80/0.8698}} & {\color{red}{30.94/0.8724}} \\
		\hline & ¡Á2 & 32.47/0.9062 & 32.52/0.9071 & 32.53/0.9065 & 32.62/0.9074 & 32.57/0.9074 & {\color{blue}{ 32.72/0.9084}} & {\color{red}{32.75/0.9088}} \\
		Set14 & ¡Á3 & 29.32/0.8200 & 29.41/0.8222 & 29.42/0.8220 & 29.48/0.8231 & 29.50/0.8237 &{\color{blue}{  29.58/0.8252 }}& {\color{red}{29.64/0.8263}} \\
		& ¡Á4 & 27.54/0.7511 & 27.62/0.7542 & 27.63/0.7541 & 27.70/0.7556 & 27.71/0.7564 &{\color{blue}{  27.76/0.7579}} & {\color{red}{27.81/0.7599}} \\
		\hline  & ¡Á2 & 31.34/0.8882 & 31.38/0.8894 & 31.38/0.8888 & 31.44/0.8897 & 31.43/0.8898 & {\color{blue}{ 31.52/0.8905 }}& {\color{red}{31.55/0.8914}} \\
		B100 & ¡Á3 & 28.37/0.7860 & 28.42/0.7879 & 28.43/0.7879 & 28.47/0.7888 & 28.49/0.7897 &{\color{blue}{  28.53/0.7910}} & {\color{red}{28.58/0.7921}} \\
		& ¡Á4 & 26.87/0.7133 & 26.93/0.7158 & 26.93/0.7157 & 26.96/0.7166 & 26.98/0.7175 & {\color{blue}{ 27.01/0.7184}} & {\color{red}{27.06/0.7203}} \\
		\hline  & ¡Á2 & 29.57/0.8961 & 29.59/0.8967 & 29.62/0.8966 & 29.72/{\color{blue}{ 0.8982}} & 29.70/0.8977 &{\color{red}{ 29.97/0.9012}} &{\color{blue}{  29.94}}/{\color{red}{0.9012}} \\
		Urban100 & ¡Á3 & 26.22/0.7978 & 26.28/0.8002 & 26.32/0.8010 & 26.38/0.8027 & 26.42/0.8038 &{\color{blue}{  26.53/0.8070}} &{\color{red}{ 26.60/0.8092 }}\\
		& ¡Á4 & 24.44/0.7187 & 24.51/0.7220 & 24.55/0.7233 & 24.58/0.7247 & 24.62/0.7265 & {\color{blue}{ 24.69/0.7288}} & {\color{red}{24.77/0.7362}} \\
		\hline
	\end{tabular}
	
	\label{table1}
\end{table*}

\begin{figure}\small
  \centering
  \includegraphics[width=0.45\textwidth]{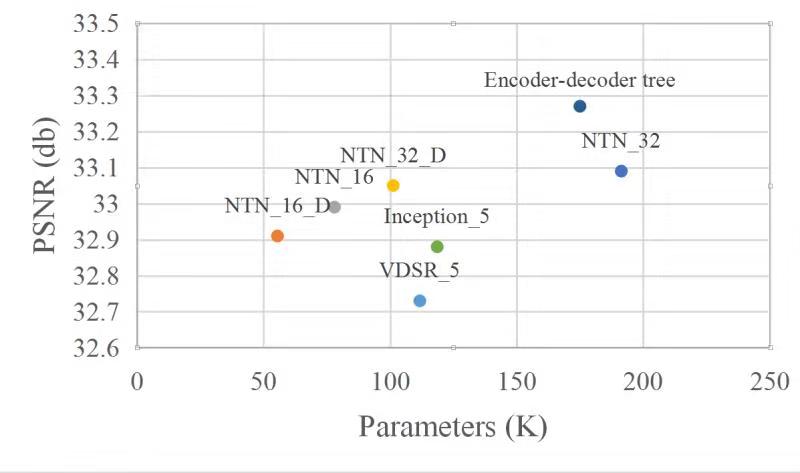}\\
  \caption{The relationship between PSNR and number of parameters. The results are evaluated on Set5 for 3¡Á enlargement.}
  \label{fig5}
\end{figure}

As shown in Fig. \ref{fig4}, the NTN for SR have seven nodes in tree structure and five layers totally. Our model consists of one tree structure and two convolutional layers. By the way, other models can also convert to NTN by adding a tree structure. The skip connection between the input image and the output image is preserved, enabling the network to converge faster \cite{he2016deep}.

\begin{figure*}\small
	\centering
	\includegraphics[width=0.95\textwidth, height=0.35\textheight]{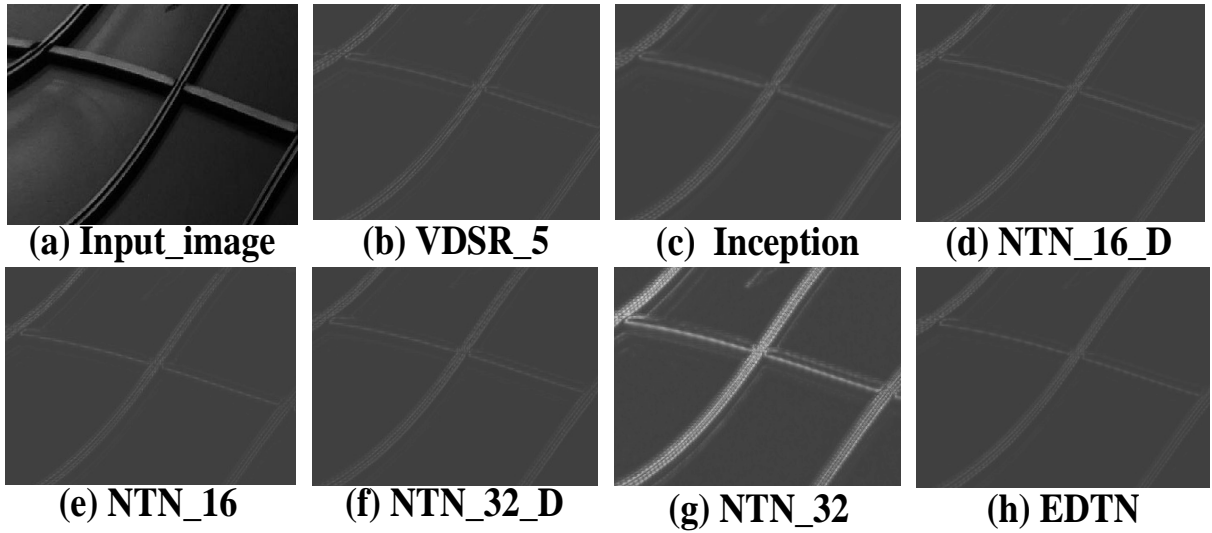}\\
	\caption{Sample of activation maps from the output of our tree structures and corresponding activation maps from VDSR\_5 and Inception.}
	\label{fig6}
\end{figure*}

We also have investigated the relationship between the number of nodes and performance. In addition, the impact of the position of the tree structure in NTN is also investiga- ted, see detail in Section \ref{Model}.

\subsection{Variant of tree}
As we all know, tree has many forms, manly determined by their branches and nodes, like binary tree, non-binary tree. When apply the tree structure in deep learning, the form of tree can be more various (caused by diversity of convolutional size). In this paper, we mainly discussed two variants of tree, which are reverse tree and encoder-decoder tree.\vspace{5mm}

\noindent\textbf{Reverse Tree Network (RTN).} Compared with the ordinary tree, of which the number of nodes in each layer is greater than or equal to the number of nodes in previous layer, the reverse tree is just the opposite.

 As Fig. \ref{fig2} (d) shows, reverse tree is equivalent to rotat-ing the ordinary tree 180 degrees. Two parent nodes share a child node, which means child nodes obtain the sum of feature from two parent nodes. Different from ordinary tree which only mixes features once, the reverse tree mixes features in each layer. Accordingly, in parameter amount, the reverse tree is slightly larger than the ordinary tree.

We found that the reverse tree did not show superiority to ordinary tree in SR. Therefore, the ordinary tree was mainly discussed in this paper. However, the application of reverse tree in other fields still worth exploring.\vspace{5mm}

\noindent\textbf{Encoder-decoder Tree Network (EDTN). } The encoder-decoder tree combines the ideology of auto-encoder \cite{zhang2017split} and tree structure. As Fig. \ref{fig2} (e) shows, the encoder-decoder tree consists one ordinary tree and one reverse tree. In specific, the ordinary tree is the encoder and reverse tree is decoder. As with auto-encoder, the ordinary tree encodes input to feature of the hidden layer and the reverse tree decodes the feature from hidden layer to output.

In this work, we only apply encoder-decoder tree to SR, of which the performance is presented in Section \ref{Model}. The application of this structure in other fields (e.g. image inpainting) is still worth exploring.

\section{ Experimental Results}

\subsection{Datasets}
In SR field, there are many data sets for training model, including: Set91 \cite{yang2010image}, 291images \cite{martin2001database}, DIV2K \cite{timofte2017ntire} and ImageNet dataset \cite{russakovsky2015imagenet}.

To verify the effectiveness of our model fast, Set91 was used as training set. Training images were split into 41 by 41 patches with stride 41, data augmentation (rotation and flip) was used to increase the number of samples. Besides, same as in VDSR, we also contain all scale images in the training set for accelerating training.

For testing, we use four datasets which are often used for SR, including: Set5 \cite{bevilacqua2012low}, Set14 \cite{zeyde2010single}, B100 \cite{timofte2014a+}, and Urban 100 \cite{huang2015single}. Specifically, Set5, Set14, and B100 consist of natural scenes, while Urban100 contains urban scenes.

\subsection{Implementation details}

In our NTN, the number of filters is set to 32 and 64 in tree structure and following convolution layers respectively, and the number of filters in the last layer is 1. All filter sizes are set to 3 except for the tree structure. In RTN, the number of filters in all convolutional layers is 64 except the first layer is 32 and last layer is 1. All convolutional layers are followed by rectified linear units (ReLu). The weights initialization follows He et al \cite{he2015delving}.

The hyperparameters setting for training are as follows: the learning rate is initialized to 0.1 for all layers and decre-ase by a factor of 10 for every 20 epochs for total 80 epochs. For optimization, we use SGD with momentum to 0.9 and weight decay to 1e-4. All experiments were conducted using Pytorch on NVIDIA TITAN X GPUs.

\subsection{Model analysis}
\label{Model}\vspace{5mm}

\noindent\textbf{Performance analysis.} To demonstrate the effectivene-ss of our tree models, we constructed a series of lightweight NTNs and its derivative network with five layers. VDSR reduced to five layers (VDSR\_5) without changing other structures was considered baseline correspondingly. In addition, another network based on Inception model was constructed for SR which contained two Inception module and five layers totally. In this experiment, seven models were conducted, including VDSR\_5, Inception, NTN\_32 (32 feature maps), NTN\_32\_D (dilated convolution), NTN\_16 (16 feature maps), NTN\_16\_D (dilated convolution), and EDTN.

 Compared with the traditional single-scale neural network represented by VDSR\_5, the multi-scale informa-tion brought by HEN effectively improved the network performance (Table \ref{table1}.). Inception model (a typical HEN) outperforms VDSR\_5 at all scales on four test datasets. Benefit from the richer feature information brought by the special structure, the NTN and its derived network have achieved better performance than Inception. For 2¡Á enlargement in Set5, NTN\_32 achieves 36.97 dB which better 0.25 dB than Inception. For 3¡Á and 4¡Á enlargement, NTN\_32 also better 0.21 dB, 0.19 dB than Inception respectively. In all models, EDTN achieved the best performance. Compared to NTN\_32, for 2¡Á, 3¡Á and 4¡Á enlargement, EDTN better 0.11 dB, 0.18 dB, 0.14dB than NTN\_32 respectively. The results show that the richer multi-scale information brought by our tree model and the more effective feature utilization form can improve the performance of the network effectively.\vspace{5mm}

\begin{figure*}[t]\small
	\centering
	\includegraphics[width=0.88\textwidth, height=0.31\textheight]{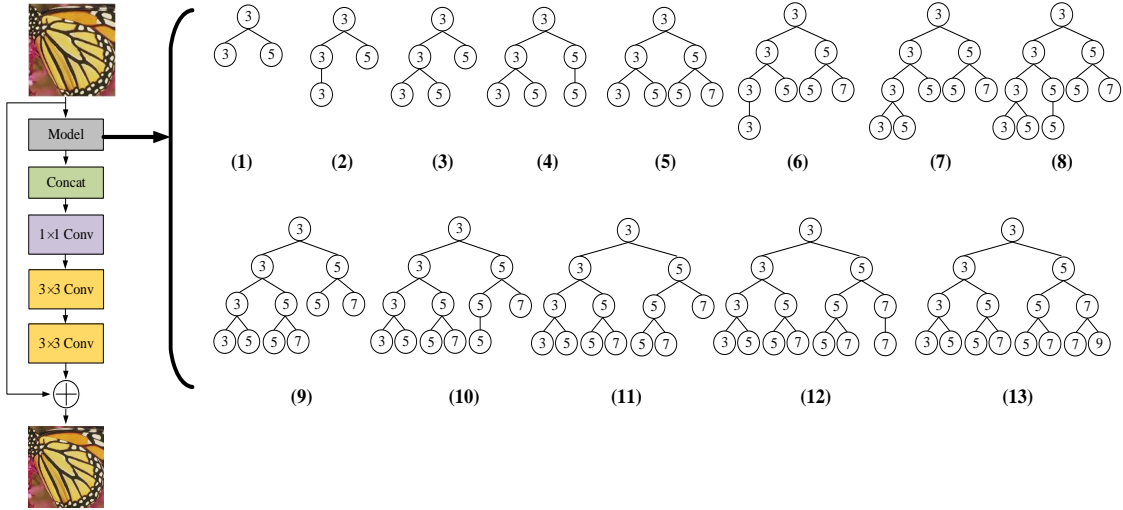}\\
	\caption{The architecture of thirteen NTNs. Thirteen models share the same network frame, as shown on the left side. The value shown in the circle is the size of the filters.}
	\label{fig7}
\end{figure*}

\noindent\textbf{Number of parameters.} Compared to VDSR\_5, the NTN\_32 are slightly larger in parameters. To reduce our model¡¯s parameters without changing the architecture, dilated convolution was introduced to replace the ordinary convolution in tree structure (e.g. NTN\_32\_D). Reducing the feature maps in tree structure (e.g. NTN\_16) was also discussed.

As Fig. \ref{fig5} shows, the NTN\_32 gets 33.09 dB which better 0.04 dB than NTN\_32\_D and NTN\_16 better 0.08 than NTN\_16\_D. It is worth noting that although the dilated convolution does not change the receptive field and structure of NTN, it will reduce the performance to certain extent. Results show that model parameters of NTN\_32\_D, NTN\_16, NTN\_16\_D are only 91\%, 70\%, 50\% of VDSR\_5.
The corresponding PSNR of these models is higher than VDSR\_5 0.32dB, 0.26dB, 0.18dB. Besides, the EDTN achieved the highest PSNR value, but its parameter amount is only 91\% of NTN\_32. These results demonstrate that our tree structure makes full use of multi-scale information to reduce parameters while improving performance.

To further compare the differences at feature level among seven models, the activation maps of NTN after feature fusion and corresponding activation maps of VDSR\_5 and Inception were extracted. Fig. \ref{fig6} shows that the activation maps of VDSR\_5 have distortions and missing details, while the introduction of multi-scale information can alleviate the distortion to a certain extent, making the activation maps more delicate.\vspace{5mm}

\begin{table}\small
	\caption{Contribution of each branch to NTN. The PSNR and CI are evaluated on Set5 for 3$¡Á$ enlargement.}
	\begin{tabular}{|c|c|c|c|c|c|c|c|c|}
		\hline
		& PSNR & Receptive field & Parameters & CI \\
		\hline Branch 1 & 32.74 & 11 & 37728 & 0.83 \\
		\hline Branch 2 & 32.89 & 13 & 54112 & 1.38 \\
		\hline Branch 3 & 32.94 & 15 & 70496 & 1.68 \\
		\hline Branch 4 & 32.99 & 17 & 95027 & 2.24 \\
		\hline NTN\_32 & 33.09 & 17 & 191328 & \\
		\hline
	\end{tabular}
	
	\label{table2}
\end{table}

\noindent\textbf{Nodes analysis.} To further investigate the relationship between the nodes and performance of tree, thirteen NTNs were designed which the nodes of tree are from three to fifteen correspondingly. The Fig. \ref{fig7} shows the architecture of all these models.

\begin{figure}\small
  \centering
  \includegraphics[width=0.45\textwidth]{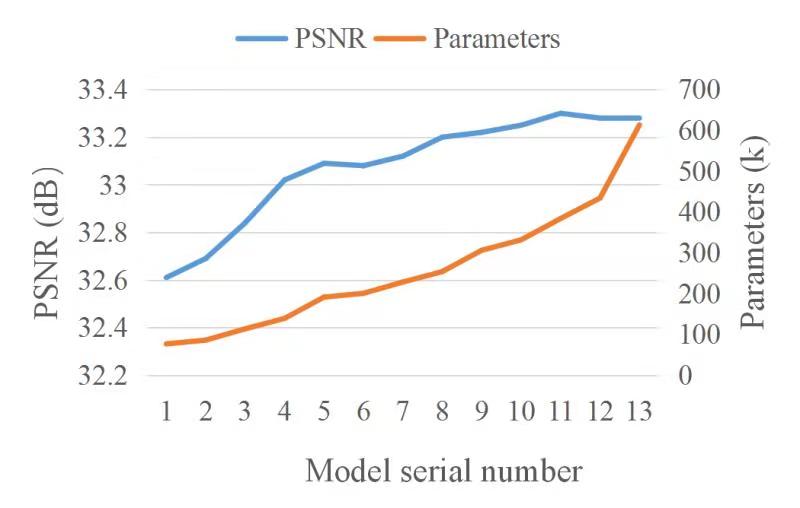}\\
  \caption{Nodes vs performance. Evaluating on Set5 for 3$¡Á$ enlargement.}
  \label{fig8}
\end{figure}

As illustrated in Fig. \ref{fig8}, the general trend is that performance improves with the number of nodes and parameters increases. However, we note that the improvement is mainly concentrated on the first five models. Subsequent model performance has not been significantly improved with the dramatic increase in the number of parameters. Thus, from the perspective of streamlined model and performance, the fifth model which has seven nodes is a good choice.\vspace{5mm}

\begin{figure}\small
	\centering
	\includegraphics[width=0.48\textwidth, height=0.20\textheight]{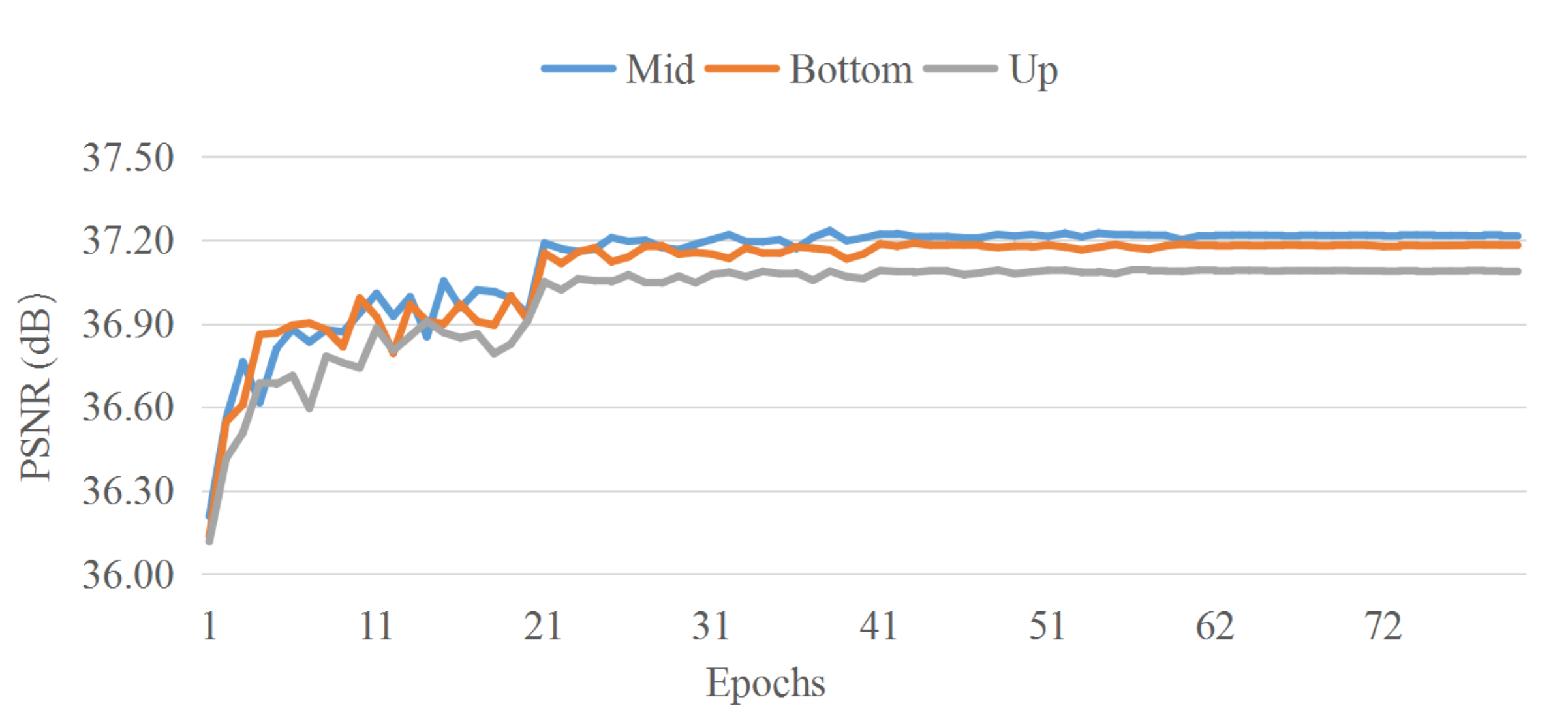}\\
	\caption{The tree position analysis in Set5 dataset for 3$¡Á$ enlargement.}
	\label{fig9}
\end{figure}

\noindent\textbf{Tree position.} Ordinary super-resolution reconstruction network consists of three parts: feature extraction (up layers), feature mapping (mid layers) and reconstruction (last layers). To explore the relationship between the position of tree in NTN (as different parts of network) and performance of model, we put the tree structure in up, middle and bottom of NTN which has 8 layers totally for comparison.

As Fig. \ref{fig9} shows, the performance difference between the tree structure in middle of NTN and bottom of NTN is not obvious, which better 0.13 dB, 0.09dB than tree structure in up of NTN for 2$¡Á$enlargement respectively. Thus, we recommend to place the tree structure in the middle or bottom of the higher performance.\vspace{5mm}

\noindent\textbf{Branch contribution analysis.} As described in Section \ref{Branch} a complete NTN can be viewed as a collection of multiple branch networks, where different branch networks contribute different degrees to overall performance. Therefore, we evaluate each branch contribution for NTN\_32.

As Table \ref{table2}. shows, the performance of branch networks is lower than that of a complete NTN network. Branch 4 has the largest receptive field and obtains the best performance. However, its parameter quantity does not have a larger increase than that of branch 3. At the same time, a certain performance improvement has been achieved at more difficult level (closer to NTN). Subjectively, it is considered that the contribution of the branch 4 to the whole network is the highest, which is consistent with the CI results, indicating that the evaluation method proposed in this paper has certain rationality.

\section{Conclusion}

In this work, we have proposed a new method to explore horizontal expansion network named NTN, which solve the common problem of HEN through the special strategy of feature sharing. We have explored some properties of NTN and presented a new form of application of the NTN structure which achieved preferable performance in SR. Besides, a new metric (CI) was defined to measure the contribution of branch networks to NTN, which is instructive to design complex tree networks in the future. The results show that the NTN and its derivative network make full use of multi-scale feature information, which effectively enhances the feature representation ability of the model.


\small
\bibliographystyle{ieee}
\bibliography{egbib}

\end{document}